%% file: acl_latex.tex
\documentclass[11pt]{article}
\usepackage{pifont}

\usepackage[preprint]{acl}

\usepackage{times}
\usepackage{latexsym}

\usepackage[T1]{fontenc}

\usepackage[utf8]{inputenc}

\usepackage{microtype}

\usepackage{inconsolata}

\usepackage{graphicx}

\usepackage{amsmath}
\usepackage{amssymb}
\usepackage{amsfonts}
\usepackage{booktabs}
\usepackage{url}
\usepackage{hyperref}
\usepackage{xurl}
\allowdisplaybreaks
\usepackage{pifont}    
\usepackage[table]{xcolor}    
\usepackage{wrapfig}
\usepackage{array}  

\usepackage[dvipsnames]{xcolor} 
\usepackage{fontawesome5}       

\usepackage{stackengine}

\usepackage{wasysym}  

\newcommand{\cmark}{\textcolor{green!60!black}{\ding{51}}}  
\newcommand{\xmark}{\textcolor{red}{\ding{55}}}             
\newcommand{\pmark}{
    \textcolor{black}
    {
        \ooalign{\ding{51}\cr\hss\raisebox{-0.7ex}{\rule{1.1ex}{0.15ex}}\hss}%
    }
}

\usepackage{xspace}
\newcommand{\OURS}{StateX\xspace}

\newcommand{\revise}[1]{\textcolor{black}{#1}}

\title{\OURS: Enhancing RNN Recall via Post-training State Expansion}

\author{Xingyu Shen$^*$, Yingfa Chen$^*$, Zhen Leng Thai, Xu Han, Zhiyuan Liu, \& Maosong Sun\\
    Department of Science and Technology, Tsinghua University, Beijing, China\\
    \texttt{xingyu-c21@mails.tsinghua.edu.cn}, \texttt{chenyingfa1999@gmail.com}
}

\begin{document}
\maketitle
\begin{abstract}

Recurrent neural networks (RNNs), such as linear attention and state-space models, have gained popularity due to their constant per-token complexity when processing long contexts. 
However, these recurrent models struggle with tasks that require accurate recall of contextual information from long contexts, because all contextual information is compressed into a fixed-size recurrent state. 
Previous studies have shown that recall ability is positively correlated with the recurrent state size, yet directly training RNNs with large recurrent states results in high training costs. 
In this paper, we introduce \OURS, a post-training framework that efficiently expands the states of pre-trained RNNs. 
For two popular classes of RNNs, linear attention and state-space models, we design post-training architectural modifications in \OURS, to scale up the state size with no or negligible increase in model parameters.
Experiments on models with up to 1.3B parameters demonstrate that \OURS~efficiently enhances the recall and in-context learning performance of RNNs without incurring high post-training costs or compromising other capabilities.

\end{abstract}

\section{Introduction}

Recently, recurrent neural networks~(RNNs), such as gated linear attention~(GLA)~\citep{yang2024gla} and Mamba2~\citep{dao2024transformers}, have shown promising performance in building large language models (LLMs). These recurrent architectures have constant per-token complexity for processing long contexts, whereas the widely-used Transformer architecture~\citep{transformer} has per-token complexity that grows linearly with the context length. Thus, RNNs are much more efficient than Transformers in processing long contexts. 

However, RNNs still underperform Transformers in certain aspects, with one of the most critical being the long-context recall ability~\citep{transformers-are-better-at-copying}. Different from Transformers, which store the representations of all contextual tokens, RNNs compress all contextual information into a fixed-size \textit{state}\footnote{Also named \textit{recurrent state} or \textit{hidden state} in various contexts. Our paper uses these two terms interchangeably.}. As a result, the recall ability of RNNs heavily depends on the capacity of this state~\citep{jelassi2024repeat,arora2024simple,yang2024gdn,chen2025stuffedmambaoversizedstates}. 
Despite the positive gains of increasing the recurrent state size, considering the increased training costs and the limited benefits in short-context scenarios and various downstream tasks, most RNNs are still trained with a small state size relative to the rest of the model. For instance, in Mamba2-1.3B and GLA-1.3B, their recurrent states are smaller than 2\% of their model sizes (details in Table~\ref{tab:state-size-and-model-size}).

\begingroup
\renewcommand{\thefootnote}{\fnsymbol{footnote}}
\footnotetext[1]{Equal contributions.}
\footnotetext[2]{The code is released at \url{https://www.github.com/THUNLP/StateX}}
\endgroup


\begin{table*}[t]
    \centering
    \footnotesize
    \begin{tabular}{l|ccc}
        \toprule
        \textbf{Method} & \textbf{Performance} & \textbf{Throughput} & \textbf{Training Cost} \\
        \midrule
        Vanilla RNNs (small states)
            & \xmark \ Poor & \cmark \ High & \cmark \ Low \\
        Training large states from scratch 
            & \pmark & \xmark \ Low & \xmark \ High \\
        Novel architectures with large states (e.g., MoM)
            & \pmark & {\xmark} \ Low & \xmark \ High \\
        \OURS (ours) 
            & \cmark \ Good & \cmark \ High & \cmark \ Low \\
        \bottomrule
    \end{tabular}
    \caption{Comparison between our work and existing approaches for increasing RNN state sizes. Vanilla RNNs underperform due to their smaller state sizes. \protect\pmark denotes methods that must be trained from scratch at a huge cost. Hence, it is difficult to set up a strictly fair comparison despite their competitive performance.}
    \label{tab:related-works}
\end{table*}

In this paper, we propose \OURS, which expands the recurrent state size while keeping training costs low and introducing almost no additional parameters. Specifically, we expand the state size of pre-trained RNNs through post-training on much less data than pre-training. Since larger recurrent states are more important for long-context models, we perform state expansion prior to continual training. The whole pipeline is illustrated in Figure~\ref{fig:pipeline}. 

The state expansion process is an architectural change and depends on the pre-trained model architecture. Therefore, for \OURS, we design two state-expansion methods, targeting two popular RNN classes: linear attention models~\citep{linear-attn,yang2024gla} and state-space models~\citep{dao2024transformers}. Additionally, we explore various parameter initialization techniques and select key layers for expansion rather than all layers, to balance model performance and adaptation efficiency. Compared to other state expansion methods that require training from scratch~(e.g., MoM~\citep{mom}), our method is simpler and can be seamlessly applied to existing effective RNN implementations and training pipelines.

We evaluate our method on public 1.3B parameter checkpoints of GLA\footnote{\url{https://huggingface.co/fla-hub/gla-1.3B-100B}} and Mamba2\footnote{\url{https://huggingface.co/AntonV/mamba2-1.3b-hf}}, two widely used RNN models, by conducting post-training on 10B tokens. Our empirical results demonstrate that, compared to the traditional two-stage method, \OURS significantly improves performance on recall-intensive tasks, in-context learning tasks, and needle-in-a-haystack~(NIAH)~\citep{ruler} tasks while maintaining performance on common-sense reasoning tasks. While using the same amount of training data as ordinary continual training, \OURS consistently yields better results: the relative accuracy gains in recall-intensive tasks are 3.36\% for GLA and 1.1\% for Mamba2, and the relative performance gains in in-context learning are 7.2\% for GLA and 1.0\% for Mamba2. Also, the average NIAH accuracies (up to 64K context length) improve from 26.0\% to 42.2\% for GLA, and from 33.2\% to 39.2\% for Mamba2.

Overall, our contributions include:

\begin{itemize}
    \item To the best of our knowledge, \OURS represents the first work that focuses on expanding the state size of RNNs through post-training.
    \item For two popular RNN variants, GLA and Mamba2, we design simple and effective state expansion techniques and training recipes for efficient post-training.
    \item We evaluate our method on public 1.3B checkpoints. Our results show consistent improvements in recall-intensive and in-context learning tasks, without sacrificing performance on common-sense reasoning benchmarks.
\end{itemize}

\section{Related Works}

In this section, we provide a brief description of RNNs and related work on expanding their state sizes. For more details about RNNs, please refer to the surveys~\citep{test-time-regression,lv2025technologieseffectivenessefficiencysurvey}.

\paragraph{Modern RNNs}

Recently, some RNN variants have shown promising results in sequence modeling. Some representative examples include 
state space models~(SSMs)~\citep{dao2024transformers,mamba}, the RWKV series~\citep{rwkv7, rwkv6, rwkv4}, linear attention (LA) models~\citep{linear-attn,retnet,yang2024gla}, and DeltaNet~\citep{yang2024gdn}. Some results have shown that these RNNs can outperform Transformers up to several billion parameters on certain language tasks, such as common-sense reasoning~\citep{waleffe2024empiricalstudymambabasedlanguage,falcon3}, and some hybrid models have scaled up to over 100B parameters and trillions of training tokens~\citep{minimax-01,kimi-linear}. 
RNNs are attractive alternatives to Transformers because their per-token complexity is constant, while Transformers' per-token complexity scales linearly with the context length. However, since Transformers cache all contextual token representations, they outperform RNNs in recalling contextual information. This is one of the reasons why RNNs have seen limited adoption.
 
\paragraph{Increasing RNN State Size}

Many previous works have investigated the influence of recurrent state size on the capabilities of RNNs. One important improvement of modern RNNs over previous works such as LSTM~\citep{lstm} and GRU~\citep{gru} is the adoption of larger matrix-valued recurrent states over smaller vector-valued states~\citep{retnet,hgrn2,linear-attn,gau}. Some later efforts focus on improving the forget mechanisms to remove unneeded information in the recurrent states, saving capacity to store more contextual information~\citep{mamba,deltanet}.
\citet{arora2024simple} provide a comprehensive comparison of the recall-throughput tradeoff of various recent RNN architectures. Although these methods show promising results, their state size is still rather small, and they lag behind Transformers in recall-intensive tasks.

\paragraph{Recent State Expansion Works}

More recently, \citet{mom} propose MoM, a new architecture that maintains a large state size but with lower computational overhead, by updating only parts of the recurrent state at each time step.


\citet{sse} is a concurrent work to ours that proposes a new row-sparse update formulation for LA, based on which they develop an architecture capable of handling large sparse states, to address the performance degradation in in-context retrieval. Another relevant concurrent work is by \citet{scaling-up-state-size}. They utilize low-rank projections to increase the state size of RNNs with small parameter overhead, resulting in considerably better recall performance. However, these architectures have not been thoroughly evaluated across different tasks and may be hard to adopt into existing codebases.

In brief, the state size is still a critical bottleneck of RNNs. Increasing the state size provides consistent performance gains for many RNN variants. However, \textbf{previous works on expanding RNN states are trained from scratch}, which is highly expensive and requires significant changes to the model architecture and implementation. This paper, to the best of our knowledge, is the first effort to \textbf{expand states through post-training}, without training from scratch. Compared to existing architectures with larger states, our method is simpler and can be seamlessly integrated into popular RNN variants such as LA and SSMs. Table~\ref{tab:related-works} shows the comparison between our work and existing works with larger states.

\section{Preliminaries}

In this section, we first provide a formulation of RNNs as well as two variants---LA and SSM (Sections~\ref{sec:preliminaries-rnn}, \ref{sec:preliminaries-gla}, and \ref{sec:preliminaries-ssm}). 
Then, we discuss how the recurrent state size influences the models' recall abilities and cost-efficiency~(Section~\ref{sec:preliminaries-influence-of-state-size}). Since we only modify the RNN block, we omit the formulation for FFN blocks.

\subsection{Recurrent Neural Networks}
\label{sec:preliminaries-rnn}

\begin{figure*}[t]
    \centering
    \includegraphics[width=\linewidth]{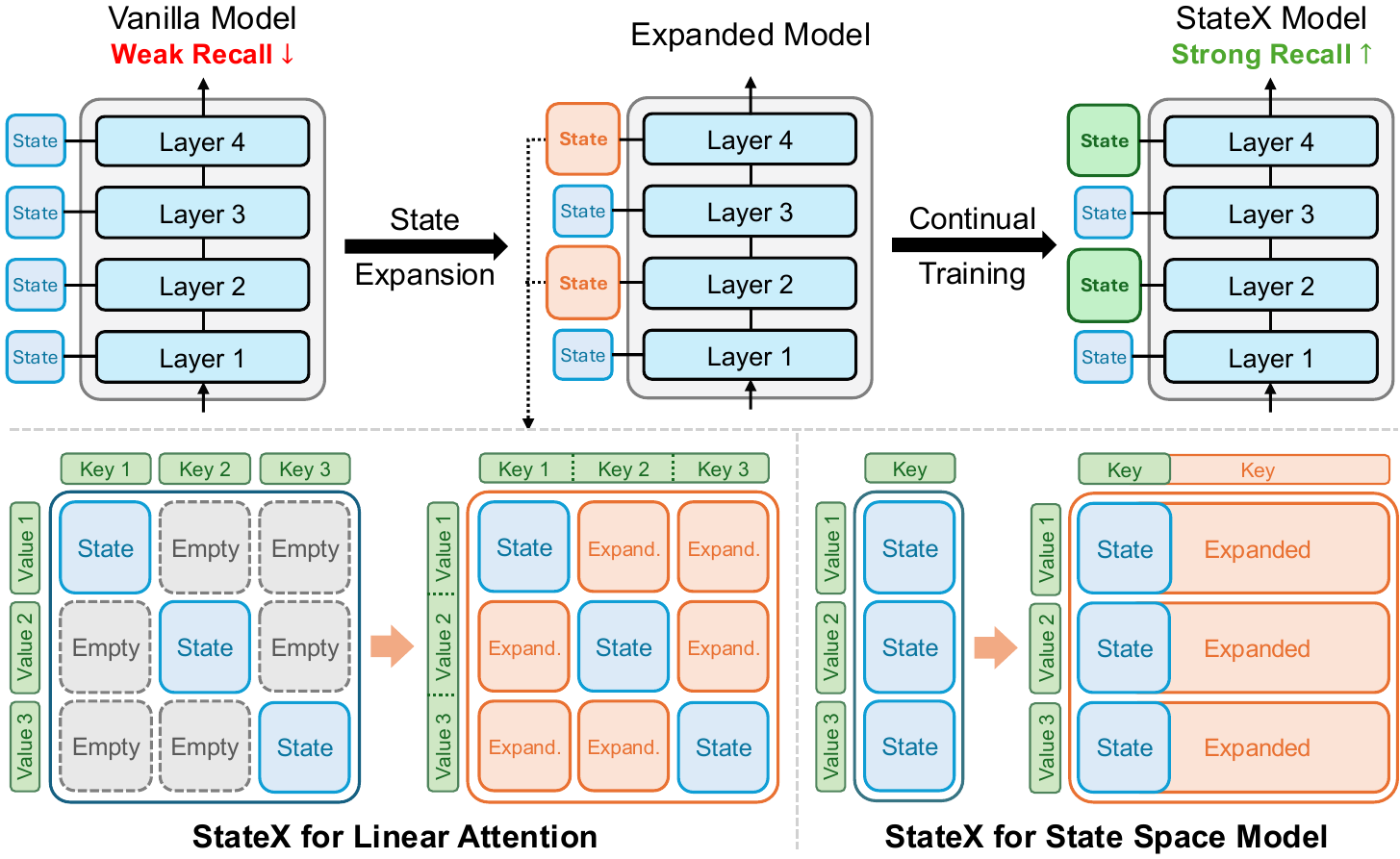}
    \caption{Illustration of the pipeline of \OURS and how \OURS expands the state size of two RNN variants (linear attention and state space models) with minimal increase in model parameters. The red parts (``Expanded State'') indicate the additional state parameters unlocked by \OURS.}
    \label{fig:pipeline}
    \label{fig:arch-modification}
\end{figure*}

In RNNs, all contextual information is stored in a fixed-size \textit{recurrent state} $\mathbf S_t$, where $t$ denotes the time step. At each time step, new information is inserted into the previous state $\mathbf S_{t-1}$ with an \textit{update rule} $f_\text{update}$, and then retrieves information from $\mathbf S_t$ with a \textit{query rule} $f_\text{query}$, which is given as
\begin{equation}
\begin{aligned}
    \mathbf S_t &= f_\text{update}(\mathbf S_{t-1}, \mathbf x_t),  \\
    \mathbf y_t &= f_\text{query}(\mathbf S_t, \mathbf x_t),
\end{aligned}
\end{equation}
where $\mathbf x_t,\mathbf y_t \in \mathbb R^{d}$ are the input and output representations at the time step $t$. In this paper, we define the \textit{state size} as the parameter count of $\mathbf S_t$.

\subsection{Linear Attention}
\label{sec:preliminaries-gla}

There are many LA variants, and we use GLA as a representative example in this study, but our method is applicable to most LA models since they have highly similar formulations.
Each GLA layer consists of $H$ heads computed in parallel, and the layer output is the sum of the heads' outputs. Each GLA head can be formulated as:\footnote{We omit some components (e.g., skip connections, normalization) for simplicity.}
\begin{equation}
\begin{aligned}
    \square_{t,h} &= f_{\square,h}(\mathbf x_t), \quad \square \in \{\mathbf{q}, \mathbf{k}, \mathbf{v}, \boldsymbol{\alpha}\},\\
    \mathbf F_{t,h} &= \text{diag}(\boldsymbol{\alpha}_{t,h})\\
    \mathbf S_{t,h} &= \mathbf F_{t,h}\mathbf S_{t-1,h} + \mathbf{k}_{t,h}^\top \mathbf{v}_{t,h} \in \mathbb{R}^{d_k \times d_v},\\
    \mathbf{y}_{t,h} &= \mathbf{q}_{t,h} \mathbf S_{t,h} \in \mathbb{R}^{d_v},
    \label{eq:gla}
\end{aligned}
\end{equation}
where $h\in\{1,\cdots,H\}$ is the head index, $d_k,d_v$ are the key and value dimensions, and $f_{\square,h}$ are differentiable functions of $\mathbf x_t$. The diagonal structure of the transition matrix $\mathbf F_{t,h}$ allows this recurrent form to be parallelized efficiently on modern GPUs~\citep{yang2024gla}. The state size in each GLA layer is $Hd_kd_v$.

\subsection{State Space Models}
\label{sec:preliminaries-ssm}

We focus on Mamba2, which is a state-of-the-art SSM. A Mamba2 layer can be formulated as:\footnote{We use attention notations ($\mathbf q_t,\mathbf k_t,\mathbf v_t$) instead of SSM notations ($x_t,B_t,C_t$) from the Mamba2 paper for simplicity and to emphasize the analogy between the two RNN variants.}
\begin{equation}
\begin{aligned}
    \Delta_{t} &= f_\Delta(\mathbf x_t \mathbf W_{\Delta,h}) \in \mathbb R,\\
    \alpha_{t,h} &= \exp(-\Delta_t \exp(A_h)) \in \mathbb R,\\
    \mathbf q_t &= \sigma_q(\mathbf x_t \mathbf W_q) \in \mathbb R^{d_k},\\
    \mathbf k_t &= \sigma_k(\mathbf x_t  \mathbf W_k) \in \mathbb R^{d_k}, \\
    \mathbf v_{t,h} &= \sigma_v(\mathbf x_t \mathbf W_{v,h}) \in \mathbb{R}^{d_v}, \\
    \mathbf S_{t,h} &= \mathbf S_{t-1,h} \alpha_{t,h} +  \Delta_{t,h}\mathbf k_t^\top \mathbf v_{t,h} \in \mathbb R^{d_k\times d_v},\\
    \mathbf y_{t,h} &= \mathbf q_t \mathbf S_{t,h}  \in \mathbb{R}^{d_v},
    \label{eq:mamba2}
\end{aligned}
\end{equation}
where $\sigma_{v},\sigma_k,\sigma_q, f_{\Delta}$ are differentiable functions, $\mathbf W_v\in\mathbb R^{d\times d_v}, \mathbf W_k, \mathbf W_q \in \mathbb R^{d\times d_k}, \mathbf W_{\Delta,h} \in \mathbb R^{d\times 1}, A_h\in\mathbb R$ are learnable parameters. $d_k$ and $d_v$ are hyperparameters and are called the \textit{state dimension} and \textit{head dimension} in the SSM literature. The state size of Mamba2 is also $Hd_k d_v$, although these hyperparameter values may differ from GLA. We provide the complete formulations of Mamba2 in Appendix~\ref{sec:appendix-mamba2-complete-formulation}.


It has been identified that Mamba2 can be viewed as a variant of GLA~\citep{yang2024gla}, where heads share the same query/key (QK) states. In this paper, we view these two variants as different variants, since this QK sharing influences state expansion, and design expansion methods specifically for GLA and Mamba2, respectively.

\subsection{Influence of State Size}
\label{sec:preliminaries-influence-of-state-size}

\paragraph{Recall Ability} 

Since all contextual information is stored in $\mathbf S_t$, the ability of RNNs to recall contextual information depends on the capacity of $\mathbf S_t$, which in turn depends on the size of $\mathbf S_t$. Extensive empirical evidence indicates a strong positive correlation between the size of the recurrent states and their performance on recall-intensive tasks~\citep{arora2024simple, gau, lact, transformers-are-better-at-copying}. These findings highlight the critical role of the state size in determining RNN recall performance, underscoring the importance of state expansion for improving recall abilities.

\paragraph{Efficiency}

The computational complexity of the token mixing component (i.e., update rule and query rule) scales linearly with the state size. Therefore, blindly increasing the state size can lead to high training and inference costs. \OURS alleviates these problems during both training and inference by expanding the states via post-training (so the model is trained with smaller states most of the time) and expanding only a subset of layers.

\section{Method}
\label{sec:method}

Our method, \OURS, involves architectural modifications that expand the RNN state sizes prior to long-context post-training to boost their recall abilities. Meanwhile, we aim to minimize the additional parameters introduced by this modification and keep the final architecture similar to the original architecture to make it easier for the modified model to adapt. An overview of the architectural modifications is illustrated in Figure~\ref{fig:arch-modification}.

In this section, we describe the state expansion recipe for two popular classes of RNNs---GLA \citep{yang2024gla} and SSM \citep{dao2024transformers} (Sections~\ref{sec:state-expansion-for-gla} and \ref{sec:state-expansion-for-ssm}).
Then, we describe parameter initialization methods after the expansion~(Section~\ref{sec:method-initialization}) and which layers to expand~(Section~\ref{sec:method-which-layers}). 

\subsection{\OURS for Linear Attention}
\label{sec:state-expansion-for-gla}

Since GLA employs a multi-head mechanism with different query, key, and value (QKV) vectors for each head, we can increase the state size by simply merging multiple heads into one larger head. This is because the state size of $H$ heads is $H\times d_k\times d_v$, and merging them into one head results in a state size of $1\times Hd_k\times Hd_v$, which is $H$ times larger. Meanwhile, no additional parameters are introduced since the total number of channels in the QKV vectors remains the same.
The effect of this change is illustrated in the left side of Figure~\ref{fig:arch-modification}. Merging GLA heads activates non-diagonal regions of the state matrix, thereby achieving larger states than the multi-head counterparts. \OURS for GLA can be formulated as:
\begin{equation}
\begin{aligned}
\mathbf{q}_t' &= \begin{bmatrix} \mathbf q_{t,1} & \cdots & \mathbf q_{t,H} \end{bmatrix}\in\mathbb R^{Hd_k}, \\
\mathbf{k}_t' &= \begin{bmatrix} \mathbf k_{t,1} & \cdots & \mathbf k_{t,H} \end{bmatrix}\in\mathbb R^{Hd_k}, \\
\mathbf{v}_t' &= \begin{bmatrix} \mathbf v_{t,1} & \cdots & \mathbf v_{t,H} \end{bmatrix}\in\mathbb R^{Hd_v}, \\
\boldsymbol{\alpha}_t' &= \begin{bmatrix} \boldsymbol{\alpha}_{t,1} & \cdots & \boldsymbol{\alpha}_{t,H} \end{bmatrix}\in\mathbb R^{Hd_k},\\
\mathbf{F}_t' &= \text{diag}(\boldsymbol{\alpha}_{t}') \in \mathbb R^{H d_k} ,\\
\mathbf S_{t}' &= \mathbf F_{t}' \mathbf{S}_{t-1}' + \mathbf {k}_{t}'^\top \mathbf {v}_{t}' \in \mathbb R^{Hd_k \times Hd_v},\\
\mathbf y_{t} &= \mathbf{q}_{t}' \mathbf {S}_{t}' \in \mathbb R^{Hd_v}.
\end{aligned}
\end{equation}
In practice, we can efficiently implement \OURS GLA by updating the values of $H,d_k,d_v$ to $\hat H = 1,\hat d_k = H d_k,\hat d_v = H d_v$. Thus, this modification can be seamlessly applied to existing GLA implementations. \OURS always merges all heads into one large head, which is motivated by the finding that single-head GLA consistently outperforms multi-head GLA (reported in Section~\ref{sec:head_ablation}).





\subsection{\OURS for SSM}
\label{sec:state-expansion-for-ssm}

The head merging method is not applicable to SSMs because there is only one query and key vector in each layer. For this RNN variant, we increase the key dimension by expanding the key and query projection layers. Specifically, we increase the hyperparameter $d_k$ (the original Mamba2 paper refers to this as the \textit{state dimension}) by appending new parameters $\left(\mathbf{\hat W}_q, \mathbf{\hat W}_k\in \mathbb R^{d\times (\hat d_k - d_k)}\right)$ to $\mathbf W_q, \mathbf W_k$, where $\hat d_k$ is the new expanded key dimension in \OURS. \OURS for SSM can be formulated as:
\begin{equation}
\begin{aligned}
\mathbf W'_q &= \begin{bmatrix} \mathbf W_q & \mathbf {\hat W}_q\end{bmatrix}  \in \mathbb R^{d \times \hat d_k},\\
\mathbf W'_k &= \begin{bmatrix} \mathbf W_k & \mathbf {\hat W}_k\end{bmatrix}  \in \mathbb R^{d \times \hat d_k},\\
\mathbf q_t' &= \sigma_q(\mathbf x_t \mathbf W_q') \in\mathbb R^{\hat d_k} ,\\
\mathbf k_t' &= \Delta_{t,h} \sigma_k(\mathbf x_t \mathbf W_k')\in\mathbb R^{\hat d_k} ,\\
\mathbf S_{t,h}' &= \mathbf S_{t-1,h}' \alpha_{t,h} + \Delta_{t,h} \mathbf { k_t'^\top}  \mathbf v_{t,h} \in \mathbb R^{\hat d_k \times d_v},\\
\mathbf y_{t,h} &= \mathbf q_{t}' \mathbf S_{t,h}' \in\mathbb R^{Hd_v}.
\end{aligned}
\label{eq:statex-mamba2}
\end{equation}
Since $\mathbf W'_k,\mathbf W'_q$ are much smaller than the other components, the increase in total parameters is less than 1\% when we use $\hat d_k = 4d_k$. This modification is illustrated by Figure~\ref{fig:arch-modification} (right). A complete formulation is given in Appendix~\ref{sec:appendix-reinitialization}.

\begin{figure*}[t]
    \includegraphics[width=0.99\linewidth]{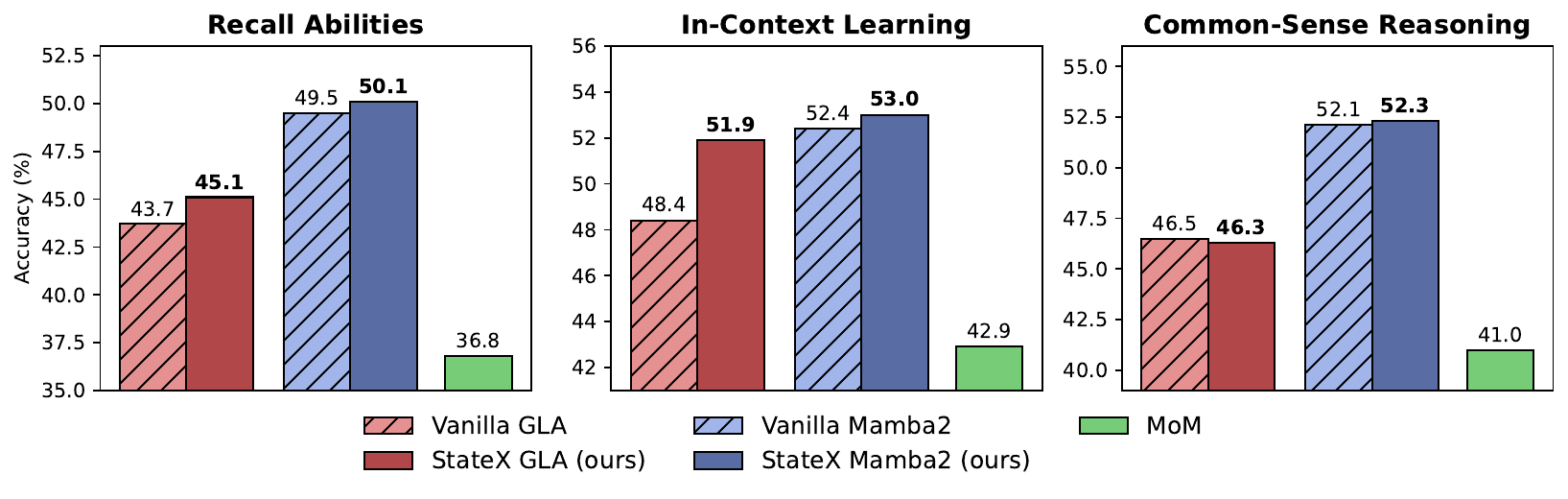}
    \caption{Comparison of the performance of vanilla RNNs (after continual training), \OURS (ours), and MoM (RNN with sparse states). \OURS improves context utilization abilities without sacrificing common-sense reasoning.}
    \label{fig:performance}
\end{figure*}

\subsection{Parameter Initialization}
\label{sec:method-initialization}

After the modification, we can inherit the parameters from the pre-trained model and initialize only the added parameters (for SSMs). However, perhaps surprisingly, we find that inheriting pre-trained parameters can be detrimental to downstream performance. Thus, we present a better parameter initialization strategy. 
An ablation study on initialization strategies is provided in Section~\ref{sec:merge}.

\paragraph{GLA Initialization}

GLA models consist of interleaving layers of GLA blocks and FFN blocks. After state expansion, we reinitialize all parameters in the GLA blocks, while FFN blocks and the embedding table inherit the pre-trained parameters.

\paragraph{SSM Initialization}

Mamba2 merges FFN blocks and the SSM mechanism into one unified layer. Motivated by the SSM literature, we only reinitialize the parameters of the SSM mechanism, which are $\mathbf W_q,\mathbf W_k$, while other modules inherit the pre-trained parameters. Further implementation details can be found in Appendix~\ref{sec:appendix-reinitialization}.

\begin{table}[t]
    \centering
    \footnotesize
    \begin{tabular}{l|cc}
        \toprule
        \textbf{Model} & \textbf{Model size} & \textbf{State size} \\
        \midrule
        \multicolumn{3}{c}{\textit{Linear Attention---GLA}} \\
        \midrule
        Vanilla GLA   & 1.37B & 12.6M  \\
        \OURS (ours)  & 1.37B & 18.9M  \\
        \midrule
        \multicolumn{3}{c}{\textit{State Space Model---Mamba2}} \\
        \midrule
        Vanilla Mamba2 & 1.34B & 25.0M \\
        \OURS (ours)   & 1.35B & 37.4M \\
        \midrule
        \multicolumn{3}{c}{\textit{RNN with Sparse State}} \\
        \midrule
        MoM            & 1.55B & 31.5M \\
        \bottomrule
    \end{tabular}
    \caption{The model and state sizes of GLA and Mamba2 variants as well as MoM.}
    \label{tab:state-size-and-model-size}
\end{table}

\subsection{How Many Layers to Expand?}
\label{sec:method-which-layers}

Expanding the states of all layers may result in a too disruptive change, making it harder for the modified model to recover from this change through post-training with limited data. It may also result in lower inference throughput due to the increase in total state size. Existing works have shown that not all layers are responsible for recalling information~\citep{bick2025understanding}. Thus, we hypothesize that only a subset of layers can benefit from a larger state. Concretely, we adopt a uniform expansion strategy by expanding one layer every $m$ consecutive layers, starting from the first layer. For both GLA and Mamba2, we set the value of $m$ to ensure that the number of expanded layers is exactly 4. In Section~\ref{sec:drop_prop}, we empirically ablate the influence of the number of expanded layers. Table~\ref{tab:state-size-and-model-size} reports the model and state sizes of each model studied in this paper. Let $N_S$ denote the state size of vanilla GLA/Mamba2, and $\hat N_S$ denote the state size of their \OURS versions, then the total state size of \OURS is $N_S \frac L m + \hat N_S(1-\frac L m)$.

\section{Experiments}

We first describe the details of the experiments (Section~\ref{sec:experimental-details}). 
Then, we present the main results of our method (Section~\ref{sec:main-result}).
We also measure the inference and training efficiency of the models (Sections~\ref{sec:inference-efficiency} and \ref{sec:training-efficiency}).
Finally, we provide ablation studies involving the choices of parameter initialization (Section~\ref{sec:merge}), the number of expanded layers (Section~\ref{sec:drop_prop}), multi-head mechanism in GLA (Section~\ref{sec:head_ablation}).
Some additional results are reported in Section~\ref{sec:niah} and \ref{sec:training-loss} to save space.

\subsection{Experimental Details}
\label{sec:experimental-details}

\paragraph{Models}

We apply \OURS to the official 1.3B checkpoints of GLA and Mamba2. For Mamba2, \OURS increases the $d_k$ hyperparameter from 128 to 512. For GLA, we merge the four existing heads from the pre-trained checkpoint into a single large head. Hence, the expanded layers in the \OURS versions have 4$\times$ larger states in both models.

\paragraph{Training Data}

All models are trained on SlimPajama~\citep{cerebras2023slimpajama}, a widely-used, high-quality, and deduplicated corpus with 627B tokens extracted from the Internet. We concatenate documents with a special token as the delimiter. Then, these concatenations are split into chunks of the specified training context length.

\paragraph{Training Configuration}

The training follows common practices in context length extension by post-training as closely as possible. Concretely, we use the cosine learning rate scheduler, with a maximum learning rate of 3e-4, and a warmup phase of 5\% of the total training steps. We use 64K context length because \citet{prolong} have shown that training with longer contexts improves recall capabilities. All models are trained with 10B tokens, with a batch size of 0.5M tokens. 

\begin{figure*}[t]
    \centering
    \includegraphics[width=0.99\linewidth]{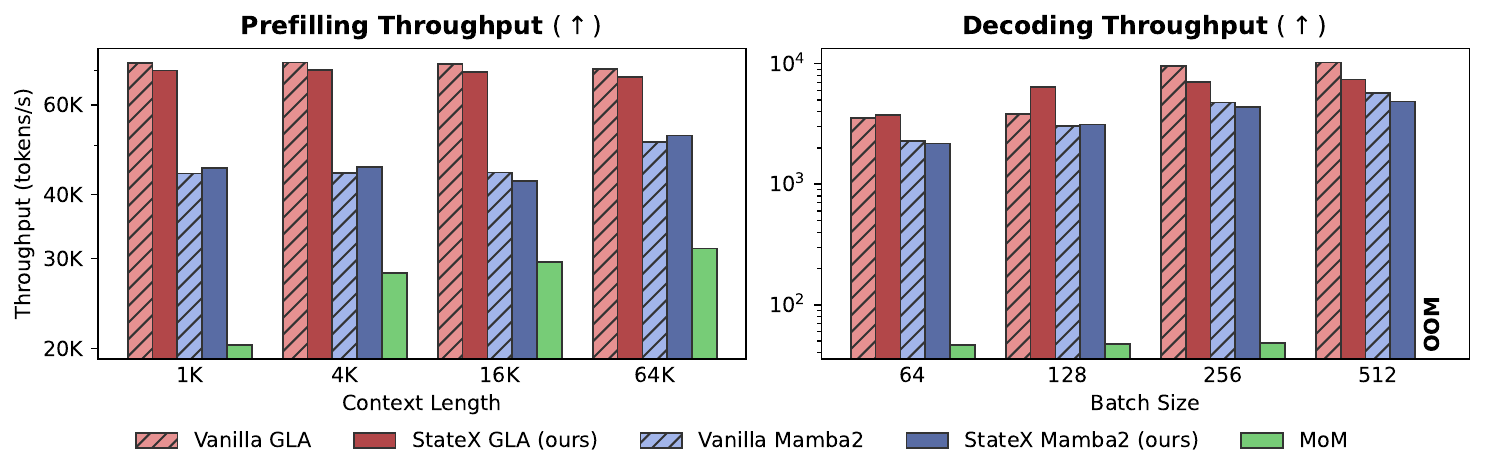}
    \caption{The inference throughput of \OURS GLA and Mamba2 versus MoM. \OURS improves the RNNs' recall while enjoying great inference speed whereas MoM is much slower in inference. OOM = Out of CUDA memory.}
    \label{fig:throughput}
\end{figure*}

\paragraph{Evaluation}

We evaluate the models' context utilization abilities with recall-intensive tasks and in-context learning (ICL). The recall-intensive tasks involve 5 popular document question-answering tasks. To assess ICL, we adopt a suite of 7 classification and 5 multiple-choice tasks selected from~\citet{min-etal-2022-rethinking}, a study that systematically evaluates ICL capabilities. The ICL prompt contains 16 demonstrations, and the performance is summarized by the mean accuracy averaged over all tasks.
Furthermore, we measure the general language processing abilities with 6 popular multiple-choice common-sense reasoning tasks. More details are given in Appendix~\ref{sec:appendix-evaluation-details}.

\paragraph{Baselines}

We mainly compare \OURS against vanilla RNNs that have undergone the continual training. The vanilla models undergo the same post-training process, but without any architectural modifications, so their state sizes remain unchanged. We also compare against MoM~\citep{mom}, which is an RNN with large states with sparse updates\footnote{As works such as \citet{sse} are closed-source and no checkpoints are released, they are excluded from our quantitative comparison.}. However, as MoM is unable to leverage pre-trained checkpoints, it is trained from scratch. For a fair comparison, throughout the entire continual training process, \OURS and each baseline are trained on the same data with the same configuration (details in Appendix~\ref{sec:appendix-baseline-settings}).

\subsection{Main Results}
\label{sec:main-result}

Figure~\ref{fig:performance} presents the scores of each model on recall-intensive, in-context learning, and common-sense reasoning tasks. The scores on each task within these three domains are given in Appendix~\ref{sec:appendix-detailed-results}. 

\paragraph{Takeaways}

\OURS considerably improves the recall and in-context learning abilities of GLA and Mamba2 over the vanilla versions. \OURS also outperforms MoM by a large margin since it can benefit from inheriting abilities from the pre-trained checkpoint while MoM has to learn from scratch.

\begin{table}[t]
    \centering
    \footnotesize
    \begin{tabular}{l|c}
        \toprule
        \textbf{Model} & \textbf{Throughput (tokens/s)} \\
        \midrule
        Vanilla GLA & \revise{129.1K} \\
        \rowcolor{blue!8}
        \OURS GLA & \revise{122.1K} \\
        \midrule
        Vanilla Mamba & \revise{108.5K} \\
        \rowcolor{blue!8}
        \OURS Mamba & \revise{104.3K} \\
        \midrule
        MoM & \revise{55.9K} \\
        \bottomrule
    \end{tabular}
    \caption{\revise{Training throughput of vanilla GLA and Mamba2, their \OURS versions and MoM. The \OURS models have a close throughput to vanilla ones, while they are roughly $2\times$ faster than MoM.}}
    \label{tab:training_throughput}
\end{table}







\subsection{Inference Efficiency Results}
\label{sec:inference-efficiency}

Here, we measure the inference throughput of the vanilla GLA and Mamba2, their \OURS versions, and MoM~\citep{mom}. The RNN mixer of each model is implemented with kernels from the widely-used Flash Linear Attention GitHub repository\footnote{\url{https://www.github.com/fla-org/flash-linear-attention}}. Inference throughput measurements are performed on a NVIDIA A800-SXM4-80GB GPU. The concrete throughput values and additional details are reported in Section~\ref{sec:appendix-efficiency}.

\paragraph{Takeaways} \OURS versions of GLA and Mamba2 almost as fast as the original models in prefilling, and slightly slower in decoding for larger batch sizes. Compared to MoM, \OURS is $1.7\times$ to $2.5\times$ faster during prefilling and $81\times$ to $147\times$ faster during decoding.

\subsection{Training Efficiency Results}
\label{sec:training-efficiency}

Table~\ref{tab:training_throughput} reports the training efficiency of vanilla RNNs, \OURS versions of them, and MoM. Training throughput is measured on a machine equipped with eight NVIDIA A800-SXM4-80GB GPUs. The training framework is implemented with the popular HuggingFace Accelerate framework with data parallelism, a common approach for single-machine, multi-GPU training.

\paragraph{Takeaways} \OURS's training speed is comparable to vanilla RNNs, and $2\times$ faster than MoM.

\subsection{Parameter Initialization Ablation}
\label{sec:merge}



Although it may seem natural to inherit pre-trained parameters, our experiments show that reinitializing the modified parameters yields better performance. For Mamba2, \OURS introduces new parameters ($\mathbf{\hat W_{q}}, \mathbf{\hat W_{k}}$) into the Q and K projections, followed by a joint reinitialization of the expanded projection matrices. In contrast, the inheritance strategy for Mamba2 preserves existing parameters while zero-initializing new introduced parameters.

\paragraph{Takeaways}

\begin{figure}[!t]
    \centering
    \includegraphics[width=\linewidth]{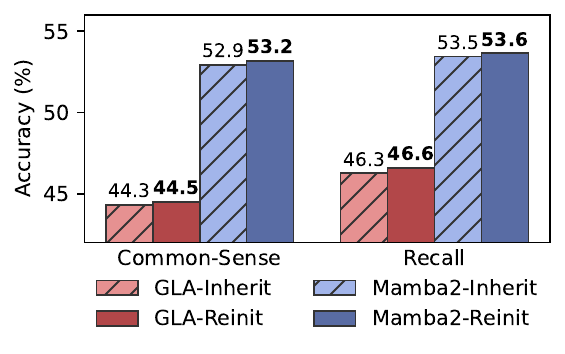}
    \caption{Model performance of reinitialization and parameter inheritance.}
    \label{fig:merge}
\end{figure}

As illustrated in Figure~\ref{fig:merge}, the model with reinitialized parameters (Reinit) consistently outperforms the one that inherits parameters (Inherit) on both common-sense reasoning and recall tasks. We hypothesize that the performance gap arises because the inherited parameters have already converged, making it difficult to effectively utilize the newly introduced channels (indicated in red in Figure~\ref{fig:arch-modification}) via post-training.

\begin{figure}[t]
    \includegraphics[width=\linewidth]{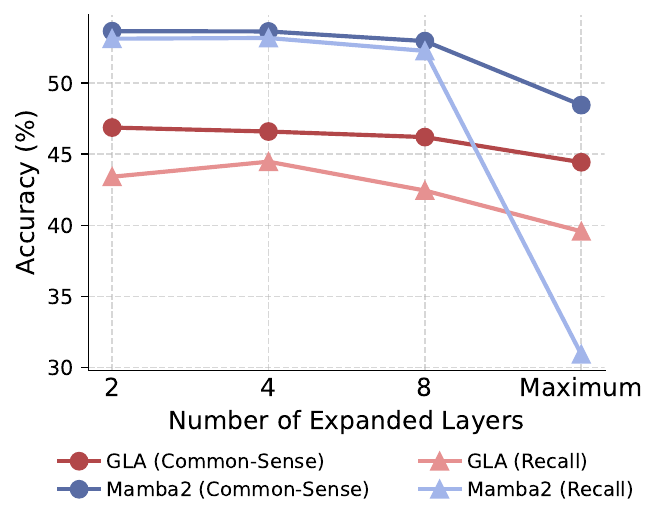}
    \caption{Model performance under varying numbers of expanded layers. Mamba2 has twice as many layers as GLA because it does not have FFN layers. "Maximum" means all layers are expanded.}
    \label{fig:drop_prop}
\end{figure}

\subsection{Best Proportion of Expanded Layers}
\label{sec:drop_prop}

As mentioned in Section~\ref{sec:method-which-layers}, it is important to balance the number of expanded layers.
To investigate this trade-off, we conducted an ablation study by varying the number of expanded layers. 

\paragraph{Takeaways}

The results, shown in Figure~\ref{fig:drop_prop}, indicate that both the GLA and Mamba2 models achieve optimal average performance when four layers are expanded (out of 24 layers and 48 layers, respectively). When too many layers are modified, the reinitialized parameters fail to converge effectively under limited continual training, leading to a sharp drop in overall performance.

\subsection{The Optimality of Single-Head GLA}
\label{sec:head_ablation}

\begin{table}[t]
    \centering
    \footnotesize
    \begin{tabular}{p{13mm}|ccc}
        \toprule
        \textbf{Head number} 
            & \textbf{CSR} $\uparrow$ 
            & \textbf{Recall} $\uparrow$ 
            & \textbf{Train loss} $\downarrow$ \\
        \midrule
        1 
            & \textbf{42.7} 
            & \textbf{26.0} 
            & \textbf{2.722} \\
        4 & 42.0 & 24.0 & 2.762 \\
        8 & 42.4 & 21.8 & 2.798 \\
        16 & 41.5 & 15.4 & 2.883 \\
        \bottomrule
    \end{tabular}
    \caption{Common-sense reasoning (CSR), recall, and training loss of GLA-340M models with different numbers of heads. Single-head GLA outperforms other configurations due to larger states.}
    \label{tab:head_comparison}
\end{table}

As mentioned in Section~\ref{sec:state-expansion-for-gla}, the multi-head mechanism in GLA significantly reduces the size of the recurrent state, which in turn leads to a degradation in model performance. This section presents an ablation study on the number of heads for GLA models trained from scratch. Each GLA model has 340M parameters and is trained on 20B tokens from the SlimPajama dataset~\citep{cerebras2023slimpajama} (see Section~\ref{sec:appendix-details-gla-heads} for more details).  Table~\ref{tab:head_comparison} reports the performance of these models on a range of common tasks. As shown, the single-head model achieves higher average scores on the benchmark tasks and converges to a lower final training loss.

\paragraph{Takeaway}

Given the same configurations and number of parameters, fewer heads allow a larger state size, leading to improved performance in common-sense reasoning, recall, and training loss.








\section{Conclusions}

We have proposed \OURS, a novel method for enhancing the recall abilities of two popular RNN variants (linear attention and state space models) by expanding the state sizes of pre-trained RNNs through post-training. Compared to training RNNs with larger state sizes from scratch, our method has much lower training costs and can be seamlessly applied to existing pre-trained models of said RNN variants. \OURS is valuable for closing the gap in the recall abilities of RNNs and Transformers. This work represents an important step toward RNNs as an efficient alternative to architectures based on softmax attention.

\section*{Limitations}

Our work focuses on pure RNN architectures, but, at the moment, hybrid architectures that combine RNN and attention layers have gained traction. The effectiveness of \OURS on hybrid architectures is yet to be validated. In addition, we have applied only \OURS to the models with 1.3B parameters. Applying \OURS to models with tens or hundreds of billions of parameters is interesting, but models with 1.3B parameters are valuable to many practitioners and researchers.

\section*{Acknowledgements}

This work is supported by the National Key Research and Development Program of China (2024YFB4505603) and National Natural Science Foundation of China (No. 62576186) and a grant from the Guoqiang Institute, Tsinghua University.

\bibliography{custom}

\newpage

\appendix

\input{appendix}



\end{document}

%% file: appendix.tex
\begin{table*}[!t]
    \centering
    \footnotesize
    \begin{tabular}{c|cccc}
        \toprule
        \textbf{Model}  
            & \textbf{Update rule} 
            & \textbf{Query rule} 
            & \textbf{State size} 
            & \textbf{StateX state size}\\
        \midrule
        GLA    
            & $\mathbf S_{t-1,h} \text{diag}(\alpha_{t,h}) + \mathbf k_{t,h}^T \mathbf v_{t,h}$ 
            & $\mathbf q_{t,h} \mathbf S_{t,h}$ 
            & $Hd_k d_v$ & $H^2d_kd_v$ \\
        Mamba2 
            & $\mathbf S_{t-1,h} \alpha_{t,h} + \Delta_{t,h}\mathbf k_t^T\mathbf v_{t,h} $ 
            & $\mathbf q_{t} \mathbf S_{t,h} + D_h\mathbf v_{t,h}$ 
            & $Hd_kd_v$ & $Hd_v d_k E$ \\
        \bottomrule
    \end{tabular}
    \caption{Overview of GLA and Mamba2, two popular RNNs with matrix-valued recurrent states. $H,P,N,d_k,d_v$ are hyperparameters of the architectures. $E$ is the expansion ratio of StateX for SSMs, which is set to 4, as mentioned in Section~\ref{sec:state-expansion-for-ssm}}
    \label{tab:update-rule-and-query-rule}
\end{table*}

\section{Formulation of Gated Linear Attention and Mamba2}
\label{sec:appendix-complete-formulation}

For completeness, we provide the complete formulation of GLA and Mamba2 in this section. By default, we use row-vector representation and bold lowercase letters refer to vector values, bold capitalized letters refer to matrix values, while non-bold symbols are scalars.

These models are trained on the next-token prediction task, which means that their input is a sequence of token IDs and their output is a sequence of probability distributions over the vocabulary $\{1,\cdots, V\}$, where $V$ is the vocabulary size.

At the beginning, each token ID is converted to a $d$-dimensional token embedding by looking up an embedding table (also called the \textit{input embeddings}) before passing to the backbone network. Let $T$ denote the sequence length. This creates a sequence of $T$ embeddings $\mathbf X^{(0)}\in\mathbb R^{T\times d}$. On the output side, the output embeddings at each position $t\in\{1,\cdots, T\}$ are converted to a probability distribution over the vocabulary via a linear layer called the \textit{language modeling head}.

\begin{table*}[t]
    \centering
    \footnotesize
    \begin{tabular}{l|cc|ccccc|c}
        \toprule
        \textbf{Model} 
            & \textbf{Params} 
            & \textbf{Total State} 
            & \textbf{SWDE} 
            & \textbf{SQuAD} 
            & \textbf{TQA} 
            & \textbf{NQ} 
            & \textbf{Drop} 
            & \textbf{Avg.} $\uparrow$ \\
        \midrule
        \multicolumn{9}{c}{\textit{Linear Attention --- GLA}} \\
        \midrule
        Vanilla  
            & 1.365B & 12.58M
            & 47.1
            & 56.8 
            & 56.0 
            & 21.9 
            & 36.5 & 43.7 \\
    
        \rowcolor{blue!8}
        \OURS (ours)
            & 1.365B & 18.87M
            & 50.3 
            & 59.1 
            & 55.0
            & 21.8
            & 39.5
            & \textbf{45.1} \\
        \midrule
        \multicolumn{9}{c}{\textit{State Space Model --- Mamba2}} \\
        \midrule
        Vanilla
            & 1.343B & 24.96M
            & 54.1 
            & 57.8
            & 63.5
            & 36.8
            & 35.4 
            & 49.5 \\
        
        \rowcolor{blue!8}
        \OURS (ours)
            & 1.350B & 37.44M
            & 56.1
            & 57.9
            & 63.6 
            & 36.4 
            & 36.3 
            & \textbf{50.1} \\
        \midrule
        \multicolumn{9}{c}{\revise{\textit{Sparse Model -- MoM}}} \\
        \midrule
        MoM~\citep{mom}
            & \revise{ 1.552B}
            & \revise{ 31.45M}
            & \revise{ 34.4 }
            & \revise{ 49.6 }
            & \revise{ 50.1 }
            & \revise{ 16.0 }
            & \revise{ 33.9 }
            & \revise{ 36.8 } \\
        \bottomrule
    \end{tabular}
    \caption{Accuracy on recall-intensive tasks with sequences truncated to a maximum of 2K tokens, as well as the model size and state size of each model. The best scores are bolded.}
    \label{tab:recall_results}
\end{table*}

In the following discussion, we denote the input and output sequences of representations for the $l$-th layer as:
\begin{equation}
\begin{aligned}
    \mathbf X^{(l)} = \begin{bmatrix}
        \mathbf x_1^{(l)} \\
        \vdots \\
        \mathbf x_T^{(l)}
    \end{bmatrix}, \mathbf Y^{(l)} = \begin{bmatrix}
        \mathbf y_1^{(l)} \\ \vdots \\ \mathbf y_T^{(l)}
    \end{bmatrix}
\end{aligned}
\end{equation}
where $T$ is the sequence length, and $\mathbf x_t^{(l)}, \mathbf y_t^{(l)}\in\mathbb R^{1\times d}$ are the input and output representations at time step $t$ for the $l$-th layer. Since the input of each layer is the output of the previous layer, we have $\mathbf x_t^{(l)} = \mathbf y_t^{(l-1)}$. We use row-representation for all vectors, unless specified.

\subsection{Gated Linear Attention}
\label{sec:appendix-gla-complete-formulation}

The entire model of GLA consists of interleaving GLA blocks and FFN blocks.
\begin{equation}
\begin{aligned}
    \mathbf Y'^{(l)} &= \text{GLA}^{(l)}\left( \mathbf X^{(l)} \right) + \mathbf X^{(l)} \\
    \mathbf Y ^{(l)} &= \text{FFN}^{(l)}\left( \mathbf Y'^{(l)} \right) + \mathbf Y'^{(l)}
\end{aligned}
\end{equation}

Each GLA block consists of multiple heads that are computed in parallel, and the block's output is the sum of the head outputs. This can be formulated as (omitting the layer index for simplicity):
\begin{align}
    \mathbf y_t = \sum_{h=1}^H \text{GLA}_h(\mathbf x_t)
\end{align}

Each head in GLA can be formulated as:
\begin{equation}
\begin{aligned}
    \boldsymbol{\square}_{t,h} &= \mathbf x_t \mathbf W_{\square,h}, \quad \square \in \{\mathbf{q}, \mathbf{k}, \mathbf{v}, \boldsymbol{\alpha}\}, \\
    \mathbf S_{t,h}  &= \text{diag}(\boldsymbol{\alpha}_{t,h}) \mathbf S_{t-1,h} + \mathbf{k}_{t,h}^\top \mathbf{v}_{t,h}, \\
    \mathbf{o}_{t,h} &= \text{Norm}(\mathbf{q}_{t,h} \mathbf S_{t,h}) \in \mathbb R^{d_v}, \\
    \mathbf{r}_t     &= \text{SILU}(\mathbf{x}_t \mathbf W_r +  \mathbf b_r) \in\mathbb R^{d_v}, \\
    \mathbf y_{t,h}  &= (\mathbf{r}_{t,h} \odot \mathbf{o}_{t,h}) \mathbf W_{o,h}^\top \in \mathbb R^{d},\\
    \mathbf y_t      &= \sum_{h=1}^H \mathbf y_{t,h}.
\end{aligned}
\end{equation}
where $\mathbf W_{q,h}, \mathbf W_{k,h}, \mathbf W_{\alpha,h} \in \mathbb R^{d\times d_k}$, $\mathbf W_{v,h}, \mathbf W_{r,h}, \mathbf W_{o,h} \mathbb R^{d\times d_v}$ are learnable parameters, \text{SILU} is an activation function, and \text{Norm} is an RMSNorm layer.

\begin{table}
    \centering
    \footnotesize
    \begin{tabular}{l|ccc}
        \toprule
        \textbf{Models} & \textbf{8-shot} $\uparrow$ & \textbf{16-shot} $\uparrow$ & \textbf{24-shot} $\uparrow$ \\
        \midrule
        \multicolumn{4}{c}{\textit{Linear Attention---GLA}} \\
        \midrule
        Vanilla & 47.3 & 49.7 & 48.4  \\
        \rowcolor{blue!8}
        StateX (ours) & \textbf{48.1} & \textbf{52.4} & \textbf{51.9} \\
        \midrule
        \multicolumn{4}{c}{\textit{State Space Models---Mamba2}} \\
        \midrule
        Vanilla & \textbf{47.7} & 49.7 & 52.4 \\

        \rowcolor{blue!8}
        StateX (ours) & 47.6 & \textbf{52.3} & \textbf{53.0}\\
        \midrule
        \multicolumn{4}{c}{\revise{\textit{Sparse Model -- MoM~\citep{mom}}}} \\
        \midrule
        MoM & 42.6 
            & 42.2
            & 42.9 \\
        \bottomrule
    \end{tabular}
    \caption{In-context learning performance of GLA and Mamba2 variants, evaluated on 12 downstream classification tasks. Higher is better.}
    \label{tab:icl-results}
\end{table}

\begin{table*}[!t]
    \centering
    \footnotesize
    \begin{tabular}{l|cccccc|c}
        \toprule
        \textbf{Model} & \textbf{PIQA} & \textbf{Hella.} & \textbf{Wino.} & \textbf{ARC-e} & \textbf{ARC-c} & \textbf{SIQA} & \textbf{Avg.} $\uparrow$ \\
        & acc $\uparrow$ & acc $\uparrow$ & acc $\uparrow$ & acc $\uparrow$ & acc $\uparrow$ & acc $\uparrow$ & \\
        \midrule
        \multicolumn{8}{c}{\textit{Linear Attention---GLA}} \\
        \midrule
        Vanilla & 69.6 & 38.2 & 54.7 & 54.5 & 22.7 & 39.6 & 46.5 \\
        \rowcolor{blue!8}
        \OURS (ours) & 69.7 & 37.1 & 54.9 & 53.9 & 22.5 & 39.9 & 46.3 \\
        \midrule
        \multicolumn{8}{c}{\textit{State Space Model---Mamba2}} \\ 
        \midrule
        Vanilla & 73.0 & 45.4 & 59.6 & 64.3 & 29.1 & 41.1 & 52.1\\
        \rowcolor{blue!8}
        \OURS (ours) & 73.6 & 45.0 & 59.9 & 64.0 & 29.6 & 41.6 & 52.3\\

        \midrule
        \multicolumn{8}{c}{\revise{\textit{Sparse Model---MoM}}} \\
        \midrule

        \revise{ \textit{MoM~\citep{mom}} }
            & \revise{ 63.3 }
            & \revise{ 30.4 }
            & \revise{ 50.8 }
            & \revise{ 45.2 }
            & \revise{ 18.8 }
            & \revise{ 37.4 }
            & \revise{ 41.0 } \\
        
        \bottomrule
    \end{tabular}
    \caption{Performance on language modeling and zero-shot common-sense reasoning.}
    \label{tab:commonsense_results}
\end{table*}

\begin{table}[!t]
    \centering
    \footnotesize
    \begin{tabular}{l|cccc}
        \toprule
        \textbf{Context Length}  &  \textbf{4K} & \textbf{8K} &    \textbf{16K} & \textbf{32K}\\
        \midrule
        \multicolumn{5}{c}{\textit{GLA---Passkey Retrieval}} \\
        \midrule
        Vanilla & 0.74 
            & 0.41 
            & 0.13 
            & 0.01 \\
        \rowcolor{blue!8}
        \OURS (ours) 
            & \textbf{0.93} 
            & \textbf{0.77} 
            & \textbf{0.34} 
            & \textbf{0.06}\\
        \midrule
        \multicolumn{5}{c}{\textit{Mamba2---NIAH-Single-2}} \\ 
        \midrule
        Vanilla 
            & 0.83 
            & 0.43 
            & 0.30 
            & \textbf{0.09} \\
        \rowcolor{blue!8}
        \OURS (ours) 
            & \textbf{0.94} 
            & \textbf{0.61} 
            & \textbf{0.32} 
            & \textbf{0.09} \\
        \bottomrule
    \end{tabular}
    \caption{Performance on retrieving specific information~(i.e., a needle) from synthetically generated long documents up to 64K tokens.}
    \label{tab:passkey}
\end{table}

\subsection{Mamba2}

\label{sec:appendix-mamba2-complete-formulation}

Mamba2 does not have FFNs and consists only of a stack of Mamba2 blocks:
\begin{align}
\mathbf Y^{(l)}=\text{Mamba2}^{(l)} \left(\mathbf X^{(l)} \right) + \mathbf X^{(l)} 
\end{align}

Mamba2 also employs a multi-head mechanism where the layer output is the sum of the head outputs (omitting the layer index for simplicity):
\begin{align}
    \text{Mamba2}(\mathbf x_t) = \sum_{h=1}^H \text{Mamba2}_h(\mathbf x_t)
\end{align}
where $H$ is the number of heads, and $h$ is the head index. Each Mamba2 head can be formulated as:
\begin{equation}
\begin{aligned}
    \mathbf v_{t,h} &= \sigma_v(\text{Conv1D}(\mathbf x_t \mathbf W_{v,h} )) \in \mathbb R^{d_v}, \\
    \mathbf k_t     &= \sigma_k(\text{Conv1D}(\mathbf x_t \mathbf W_k     )) \in \mathbb R^{d_k}, \\
    \mathbf q_t     &= \sigma_q(\text{Conv1D}(\mathbf x_t \mathbf W_q     )) \in \mathbb R^{d_k}, \\
    \Delta_{t,h}    &= \text{SILU}(\mathbf x_t \mathbf W_{\Delta, h} + \mathbf b_{\Delta, h}) \in \mathbb R, \\
    \alpha_{t,h}    &= \exp(-\Delta_t \exp(A_h)) \in \mathbb R,\\
    \mathbf S_{t,h} &= \mathbf S_{t-1,h} \alpha_{t,h} + \Delta_{t,h} \mathbf k_t^\top \mathbf v_{t,h} \in \mathbb R^{d_k \times d_v},\\
    \mathbf o_{t,h} &= \mathbf q_t \mathbf S_{t,h} + D_{h}\mathbf v_{t,h} \in \mathbb{R}^{d_v},\\
    \mathbf z_{t,h} &= \text{SILU}(\mathbf x_t \mathbf W_{z,h}) \in \mathbb R^{d_v},\\
    \mathbf y_{t,h} &= \text{Norm}(\mathbf o_{t,h} \odot \mathbf z_{t,h}) \mathbf W_{o,h}^\top \in \mathbb R^{d},\\
    \mathbf y_t     &= \sum_{h=1}^H \mathbf y_{t,h}.
\end{aligned}
\end{equation}
where $\text{Conv1D}$ is a per-channel 1-dimensional convolutional layer with a kernel size of 4, $\mathbf W_{\Delta,h}\in\mathbb R^{d\times 1}, \mathbf b_{\Delta,h},A_h \in \mathbb R, \mathbf W_{v,h}, \mathbf W_{z,h}, \mathbf W_{o,h} \in\mathbb R^{d\times d_v}, \mathbf W_{k}, \mathbf W_{q}\in \mathbb R^{d\times d_k}$ are learnable parameters.

\subsection{Update Rule and Query Rule}

Central to recurrent architectures are the update rule and query rule (described in Section~\ref{sec:preliminaries-rnn}), which dictate how the architecture models inter-token dependencies. Table~\ref{tab:update-rule-and-query-rule} shows the update rule and query rule of GLA and Mamba2.

\subsection{Details of Parameter Reinitialization}
\label{sec:appendix-reinitialization}

In the case of GLA, we reinitialize all parameters within the GLA block, including its normalization layer. For Mamba2, \OURS reinitializes all parameters of $A_h, \mathbf W_k', \mathbf W_q'$. Moreover, we reinitialize the bias term of the projection layer for the discretization term ($\Delta_{t,h}$), which is called \texttt{dt\_bias} in the official implementation of Mamba2\footnote{\url{https://github.com/state-spaces/mamba/blob/main/mamba_ssm/modules/mamba2.py}}.

\section{More Results}

\subsection{Detailed Results on Recall, ICL, and Common-Sense Reasoning}
\label{sec:appendix-detailed-results}

Tables~\ref{tab:recall_results}, \ref{tab:icl-results}, and \ref{tab:commonsense_results} reports the details results of the models studied in this paper in each subtask measuring recall abilities, in-context learning, and common-sense reasoning.

\subsection{Improvement on Long-Context Retrieval}
\label{sec:niah}

The recall-intensive tasks we used in Section~\ref{sec:main-result} contain mostly sequences with fewer than 4K tokens. To evaluate the models' abilities to retrieve information from longer contexts, we use the popular NIAH task~\citep{ruler}. Due to differences in the recall abilities between the GLA and Mamba2, we evaluate them using NIAH tasks of varying difficulty to avoid score saturation and preserve discriminative resolution. For the GLA model, we employed the simpler passkey retrieval task from $\infty$Bench~\citep{infinitebench}, which involves retrieving a single 5-digit passkey from long documents consisting of repeated text. For Mamba2, we use the more challenging NIAH-Single-2 task from RULER~\citep{ruler}, where a 7-digit passkey is embedded in a semantically meaningful, non-repetitive distractor content. More details can be found in Appendix~\ref{sec:appendix-passkey-details}.

\paragraph{Results} Table~\ref{tab:passkey} reports the models' performances in NIAH. It shows that, by unlocking a larger state size, \OURS significantly improves the model's recall performance in long contexts.

\subsection{More Efficiency Results}
\label{sec:appendix-efficiency}

Tables~\ref{tab:inference_prefilling} and \ref{tab:decoding_throughput_large} show the detailed inference throughput results of various models studied in this paper.

\begin{table}[ht]
    \centering
    \footnotesize
    \begin{tabular}{l|cccc|c}
        \toprule
        \textbf{\revise{Ctx. len.}} & \textbf{\revise{1K}} & \textbf{\revise{4K}} & \textbf{\revise{16K}} & \textbf{\revise{64K}} & \textbf{\revise{Avg.}} \revise{$\uparrow$} \\
        \midrule
        \multicolumn{6}{c}{\revise{\textit{Linear Attention --- GLA}}} \\
        \midrule
        \revise{GLA} & \revise{72.5K} & \revise{72.6K} & \revise{72.2K} & \revise{70.4K} & \revise{71.9K} \\
        \rowcolor{blue!8}
        \revise{---StateX} & \revise{70.0K} & \revise{70.1K} & \revise{69.5K} & \revise{68.0K} & \revise{69.4K} \\
        \midrule
        \multicolumn{6}{c}{\revise{\textit{State Space Model --- Mamba}}} \\ 
        \midrule
        \revise{Mamba2} & \revise{44.0K} & \revise{44.1K} & \revise{44.2K} & \revise{50.7K} & \revise{45.7K} \\
        \rowcolor{blue!8}
        \revise{---StateX} & \revise{45.1K} & \revise{45.3K} & \revise{42.6K} & \revise{52.2K} & \revise{46.3K} \\
        \midrule
        \multicolumn{6}{c}{\revise{\textit{RNN with Sparse States --- MoM}}} \\
        \midrule
        \revise{MoM} & \revise{20.3K} & \revise{28.1K} & \revise{29.5K} & \revise{31.4K} & \revise{27.3K} \\
        \bottomrule
    \end{tabular}
    \caption{\revise{Prefilling throughput (tokens/s) of various models, measured with different context lengths.}}
    \label{tab:inference_prefilling}
\end{table}

\begin{table}[ht]
    \centering
    \footnotesize
    \begin{tabular}{l|cccc|c}
        \toprule
        \textbf{\revise{BSZ}} & \textbf{\revise{64}} & \textbf{\revise{128}} & \textbf{\revise{256}} & \textbf{\revise{512}} & \textbf{\revise{Avg.}} \revise{$\uparrow$} \\
        \midrule
        \multicolumn{6}{c}{\revise{\textit{Linear Attention---GLA}}} \\
        \midrule
        \revise{GLA} & \revise{3548} & \revise{3815} & \revise{9595} & \revise{10226} & \revise{6796} \\
        \rowcolor{blue!8}
        \revise{--StateX} & \revise{3770} & \revise{6371} & \revise{7083} & \revise{7395} & \revise{6155} \\
        \midrule
        \multicolumn{6}{c}{\revise{\textit{State Space Model---Mamba}}} \\ 
        \midrule
        \revise{Mamba} & \revise{2276} & \revise{3034} & \revise{4755} & \revise{5730} & \revise{3949} \\
        \rowcolor{blue!8}
        \revise{--StateX} & \revise{2174} & \revise{3121} & \revise{4351} & \revise{4837} & \revise{3620} \\
        \midrule
        \multicolumn{6}{c}{\revise{\textit{RNN with Sparse State---MoM}}} \\
        \midrule
        MoM & \revise{46.3} & \revise{47.5} & \revise{48.2} & \revise{\textcolor{red}{OOM}} & \revise{47.3} \\
        \bottomrule
    \end{tabular}
    \caption{\revise{Decoding throughput (tokens/s) of various models, measured with different batch sizes.}}
    \label{tab:decoding_throughput_large}
\end{table}

\subsection{Training Loss}
\label{sec:training-loss}

\begin{figure}[t]
    \includegraphics[width=\linewidth]{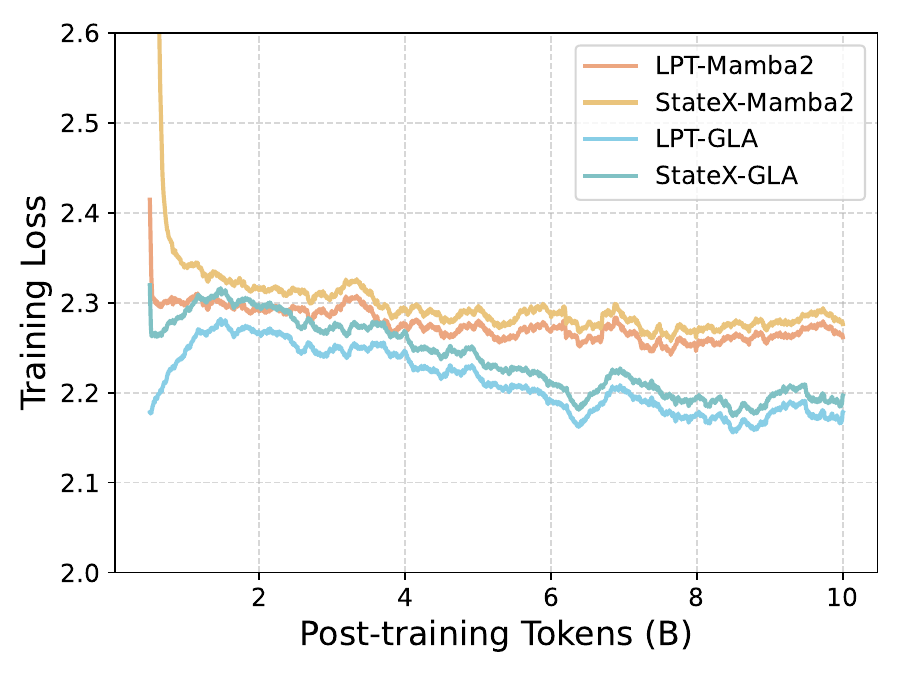}
    \caption{Post-training loss (on SlimPajama) of vanilla models and expanded models. GLA has lower loss as it is pre-trained on SlimPajama while Mamba2 is pre-trained on Pile. LPT means the vanilla models undergoing continual training.}
    \label{fig:training_loss}
\end{figure}

We also tracked the training loss curves of models trained with standard continual training and with \OURS. Figure~\ref{fig:training_loss} shows the loss curves for both GLA and Mamba2. The former has generally lower loss because it was pre-trained on SlimPajama, while Mamba2 was not.
Notably, the \OURS models have a higher initial training loss due to the architectural change, but quickly close the gap. Interestingly, although their final training loss is slightly higher than the continual training counterparts, they achieve better performance on downstream tasks.

\section{Experiment Details}

\subsection{Baselines}
\label{sec:appendix-baseline-settings}

The vanilla models are obtained through continual training on the data mentioned before. In contrast, the \OURS models incorporate an additional state expansion step before continual training. MoM~\citep{mom}, a representative sparse state model, is trained from scratch and has undergone the same amount of data throughout continual training, as there are no available pre-trained checkpoints for these novel architectures with large states.

\subsection{Evaluation}
\label{sec:appendix-evaluation-details}

We configure the evaluation tasks using the lm-evaluation-harness framework~\cite{eval-harness}. A set of widely adopted benchmark tasks is selected to assess the models' capabilities in common-sense reasoning and information recall. For the common-sense and recall tasks, we adopt \textit{accuracy} (not \textit{normalized accuracy}) and \textit{contains} as the respective evaluation metrics. \textit{Accuracy} directly reflects the correctness of the common-sense task results, while \textit{contains} measures the proportion of recall task outputs that include the passkey. Notably, for tasks related to recall ability, we adopt the ``Just Read Twice'' prompt from \citet{arora2024just}, which is also used in \citet{yang2024gla} and \citet{yang2024gdn}, given that all models under evaluation are based on recurrent architectures.

\begin{table*}[t]
    \centering
    \begin{tabular}{>{\raggedright\arraybackslash}p{0.22\textwidth} p{0.7\textwidth}}
        \toprule
        \textbf{Passkey Retrieval ($\infty$Bench)} & 
        \underline{\textbf{Task Template:}} \\
        & \textcolor{lightgray}{The grass is green. The sky is blue. The sun is yellow. Here we go. There and back again.} \\
        & \textcolor{lightgray}{......} \\
        & The pass key is \textcolor{orange}{\{number\}}. Remember it. \textcolor{orange}{\{number\}} is the pass key. \\
        & \textcolor{lightgray}{......} \\
        & \textcolor{lightgray}{The grass is green. The sky is blue. The sun is yellow. Here we go. There and back again.} \\
        & \\
        & \underline{\textbf{Task Answer Prefix:}} \\
        & What is the pass key? The pass key is \\
        \midrule
        \textbf{NIAH-Single-2 (RULER)} & 
        \underline{\textbf{Task Template:}} \\
        & Some special magic numbers are hidden within the following text. Make sure to memorize it. I will quiz you about the numbers afterwards. \\
        & \textcolor{lightgray}{Paul Graham Essays.} \\
        & \textcolor{lightgray}{......} One of the special magic numbers for \textcolor{blue}{\{word\}} is: \textcolor{orange}{\{number\}}. \textcolor{lightgray}{......} \\
        & What is the special magic number for \textcolor{blue}{\{word\}} mentioned in the provided text? \\
        & \\
        & \underline{\textbf{Task Answer Prefix:}} \\
        & The special magic number for \textcolor{blue}{\{word\}} mentioned in the provided text is \\
        \bottomrule
    \end{tabular}
    \caption{The prompt templates of the NIAH tasks used to evaluate the models in retrieving information from long contexts.}
    \label{tab:passkey_setting}
\end{table*}

\subsection{In-Context Learning Evaluation}

For the in-context learning (ICL) evaluation, we follow the setup introduced by~\citet{min-etal-2022-rethinking}, which systematically benchmarks ICL capabilities across classification and multiple-choice tasks. Our evaluation adopts the same protocol, but we also evaluate with a different number of in-context demonstrations for comprehensiveness.

The tasks that were used for evaluation are:
\begin{itemize}
    \item \texttt{commonsense\_qa}
    \item \texttt{ai2\_arc}
    \item \texttt{superglue-copa}
    \item \texttt{superglue-cb}
    \item \texttt{glue-mrpc}
    \item \texttt{glue-sst2}
    \item \texttt{glue-qqp}
    \item \texttt{glue-cola}
    \item \texttt{superglue-rte}
    \item \texttt{superglue-wic}
    \item \texttt{codah}
    \item \texttt{dream}
\end{itemize}

\subsection{Needle-in-a-Haystack Tasks}
\label{sec:appendix-passkey-details}

As mentioned in the previous section, we design two passkey retrieval tasks with varying levels of difficulty. The specific noise configurations and prompt templates used in each task are detailed in Table~\ref{tab:passkey_setting}. We use 5-digit passkeys in Passkey Retrieval and 7-digit passkeys in NIAH-Single-2. For each unique test length, the task will be tested on 256 randomly generated examples to ensure the consistency of the results.

    

\subsection{More Details: Ablation Study on the Number of GLA Heads}
\label{sec:appendix-details-gla-heads}

The training procedure for these models follows common language model pre-training practices as closely as possible. The model is trained on 20B tokens from SlimPajama, with a 0.5M tokens per batch, and a sequence length of 4k. We employ a cosine learning rate scheduler with an initial learning rate of 3e-4 and no specified minimum learning rate. All models consist of 340 million parameters and comprise 24 layers, each with an identical hidden state dimension. The only architectural difference lies in the number of attention heads: the single-head model uses one head with a dimensionality of 512, while the four-head model uses four heads, each with a dimensionality of 128, and so on, following the same principle.

\section{The Use of Large Language Models}

Large language models (LLMs) were used to quality-check the final draft, but we never explicitly instruct LLMs to write any parts of this paper.

%% file: custom.bib
@misc{kimi-linear,
      title={Kimi Linear: An Expressive, Efficient Attention Architecture}, 
      author={Team Kimi and Yu Zhang and Zongyu Lin and Xingcheng Yao and Jiaxi Hu and Fanqing Meng and Chengyin Liu and Xin Men and Songlin Yang and Zhiyuan Li and Wentao Li and Enzhe Lu and Weizhou Liu and Yanru Chen and Weixin Xu and Longhui Yu and Yejie Wang and Yu Fan and Longguang Zhong and Enming Yuan and Dehao Zhang and Yizhi Zhang and T. Y. Liu and Haiming Wang and Shengjun Fang and Weiran He and Shaowei Liu and Yiwei Li and Jianlin Su and Jiezhong Qiu and Bo Pang and Junjie Yan and Zhejun Jiang and Weixiao Huang and Bohong Yin and Jiacheng You and Chu Wei and Zhengtao Wang and Chao Hong and Yutian Chen and Guanduo Chen and Yucheng Wang and Huabin Zheng and Feng Wang and Yibo Liu and Mengnan Dong and Zheng Zhang and Siyuan Pan and Wenhao Wu and Yuhao Wu and Longyu Guan and Jiawen Tao and Guohong Fu and Xinran Xu and Yuzhi Wang and Guokun Lai and Yuxin Wu and Xinyu Zhou and Zhilin Yang and Yulun Du},
      year={2025},
      eprint={2510.26692},
      archivePrefix={arXiv},
      primaryClass={cs.CL},
      url={https://arxiv.org/abs/2510.26692}, 
}

@inproceedings{prolong,
    title = "How to Train Long-Context Language Models (Effectively)",
    author = "Gao, Tianyu  and
      Wettig, Alexander  and
      Yen, Howard  and
      Chen, Danqi",
    editor = "Che, Wanxiang  and
      Nabende, Joyce  and
      Shutova, Ekaterina  and
      Pilehvar, Mohammad Taher",
    booktitle = "Proceedings of the 63rd Annual Meeting of the Association for Computational Linguistics (Volume 1: Long Papers)",
    month = jul,
    year = "2025",
    address = "Vienna, Austria",
    publisher = "Association for Computational Linguistics",
    url = "https://aclanthology.org/2025.acl-long.366/",
    doi = "10.18653/v1/2025.acl-long.366",
    pages = "7376--7399",
    ISBN = "979-8-89176-251-0",
}

@inproceedings{arora2024simple,
  title={Simple linear attention language models balance the recall-throughput tradeoff},
  author={Arora, Simran and Eyuboglu, Sabri and Zhang, Michael and Timalsina, Aman and Alberti, Silas and Zou, James and Rudra, Atri and R{\'e}, Christopher},
  booktitle={Proceedings of the 41st International Conference on Machine Learning},
  pages={1763--1840},
  year={2024}
}

@article{lstm,
  author={Hochreiter, Sepp and Schmidhuber, Jürgen},
  journal={Neural Computation}, 
  title={Long Short-Term Memory}, 
  year={1997},
  volume={9},
  number={8},
  pages={1735-1780},
}

@inproceedings{yang2024gla,
  title={Gated Linear Attention Transformers with Hardware-Efficient Training},
  author={Yang, Songlin and Wang, Bailin and Shen, Yikang and Panda, Rameswar and Kim, Yoon},
  booktitle={International Conference on Machine Learning},
  pages={56501--56523},
  year={2024},
}

@misc{gru,
      title={On the Properties of Neural Machine Translation: Encoder-Decoder Approaches}, 
      author={Kyunghyun Cho and Bart van Merrienboer and Dzmitry Bahdanau and Yoshua Bengio},
      year={2014},
      eprint={1409.1259},
      archivePrefix={arXiv},
      primaryClass={cs.CL},
      url={https://arxiv.org/abs/1409.1259}, 
}

@inproceedings{dao2024transformers,
  title={Transformers are SSMs: Generalized Models and Efficient Algorithms Through Structured State Space Duality},
  author={Dao, Tri and Gu, Albert},
  booktitle={International Conference on Machine Learning},
  pages={10041--10071},
  year={2024},
  organization={PMLR}
}

@misc{falcon3,
    title = {The Falcon 3 Family of Open Models},
    url = {https://huggingface.co/blog/falcon3},
    author = {Falcon-LLM Team},
    month = {December},
    year = {2024}
}

@misc{test-time-regression,
      title={Test-time regression: a unifying framework for designing sequence models with associative memory}, 
      author={Ke Alexander Wang and Jiaxin Shi and Emily B. Fox},
      year={2025},
      eprint={2501.12352},
      archivePrefix={arXiv},
      primaryClass={cs.LG},
      url={https://arxiv.org/abs/2501.12352}, 
}

@misc{minimax-01,
      title={MiniMax-01: Scaling Foundation Models with Lightning Attention}, 
      author={MiniMax and Aonian Li and Bangwei Gong and Bo Yang and Boji Shan and Chang Liu and Cheng Zhu and Chunhao Zhang and Congchao Guo and Da Chen and Dong Li and Enwei Jiao and Gengxin Li and Guojun Zhang and Haohai Sun and Houze Dong and Jiadai Zhu and Jiaqi Zhuang and Jiayuan Song and Jin Zhu and Jingtao Han and Jingyang Li and Junbin Xie and Junhao Xu and Junjie Yan and Kaishun Zhang and Kecheng Xiao and Kexi Kang and Le Han and Leyang Wang and Lianfei Yu and Liheng Feng and Lin Zheng and Linbo Chai and Long Xing and Meizhi Ju and Mingyuan Chi and Mozhi Zhang and Peikai Huang and Pengcheng Niu and Pengfei Li and Pengyu Zhao and Qi Yang and Qidi Xu and Qiexiang Wang and Qin Wang and Qiuhui Li and Ruitao Leng and Shengmin Shi and Shuqi Yu and Sichen Li and Songquan Zhu and Tao Huang and Tianrun Liang and Weigao Sun and Weixuan Sun and Weiyu Cheng and Wenkai Li and Xiangjun Song and Xiao Su and Xiaodong Han and Xinjie Zhang and Xinzhu Hou and Xu Min and Xun Zou and Xuyang Shen and Yan Gong and Yingjie Zhu and Yipeng Zhou and Yiran Zhong and Yongyi Hu and Yuanxiang Fan and Yue Yu and Yufeng Yang and Yuhao Li and Yunan Huang and Yunji Li and Yunpeng Huang and Yunzhi Xu and Yuxin Mao and Zehan Li and Zekang Li and Zewei Tao and Zewen Ying and Zhaoyang Cong and Zhen Qin and Zhenhua Fan and Zhihang Yu and Zhuo Jiang and Zijia Wu},
      year={2025},
      eprint={2501.08313},
      archivePrefix={arXiv},
      primaryClass={cs.CL},
      url={https://arxiv.org/abs/2501.08313}, 
}

@misc{lv2025technologieseffectivenessefficiencysurvey,
      title={Technologies on Effectiveness and Efficiency: A Survey of State Spaces Models}, 
      author={Xingtai Lv and Youbang Sun and Kaiyan Zhang and Shang Qu and Xuekai Zhu and Yuchen Fan and Yi Wu and Ermo Hua and Xinwei Long and Ning Ding and Bowen Zhou},
      year={2025},
      eprint={2503.11224},
      archivePrefix={arXiv},
      primaryClass={cs.LG},
      url={https://arxiv.org/abs/2503.11224}, 
}

@inproceedings{jelassi2024repeat,
  title={Repeat After Me: Transformers are Better than State Space Models at Copying},
  author={Jelassi, Samy and Brandfonbrener, David and Kakade, Sham M and Malach, Eran},
  booktitle={International Conference on Machine Learning},
  pages={21502--21521},
  year={2024},
  organization={PMLR}
}

@misc{transformers-are-better-at-copying,
      title={Repeat After Me: Transformers are Better than State Space Models at Copying}, 
      author={Samy Jelassi and David Brandfonbrener and Sham M. Kakade and Eran Malach},
      year={2024},
      eprint={2402.01032},
      archivePrefix={arXiv},
      primaryClass={cs.LG},
      url={https://arxiv.org/abs/2402.01032}, 
}

@misc{yang2024gdn,
      title={Gated Delta Networks: Improving Mamba2 with Delta Rule}, 
      author={Songlin Yang and Jan Kautz and Ali Hatamizadeh},
      year={2025},
      eprint={2412.06464},
      archivePrefix={arXiv},
      primaryClass={cs.CL},
      url={https://arxiv.org/abs/2412.06464}, 
}

@misc{cerebras2023slimpajama,
    author = {Soboleva, Daria and Al-Khateeb, Faisal and Myers, Robert and Steeves, Jacob R and Hestness, Joel and Dey, Nolan},
    title = {{SlimPajama: A 627B token cleaned and deduplicated version of RedPajama}},
    year = 2023,
    howpublished = {\url{https://cerebras.ai/blog/slimpajama-a-627b-token-cleaned-and-deduplicated-version-of-redpajama}},
    url = {https://huggingface.co/datasets/cerebras/SlimPajama-627B},
}

@misc{transformer,
      title={Attention Is All You Need}, 
      author={Ashish Vaswani and Noam Shazeer and Niki Parmar and Jakob Uszkoreit and Llion Jones and Aidan N. Gomez and Lukasz Kaiser and Illia Polosukhin},
      year={2023},
      eprint={1706.03762},
      archivePrefix={arXiv},
      primaryClass={cs.CL},
      url={https://arxiv.org/abs/1706.03762}, 
}

@misc{chen2025stuffedmambaoversizedstates,
      title={Stuffed Mamba: Oversized States Lead to the Inability to Forget}, 
      author={Yingfa Chen and Xinrong Zhang and Shengding Hu and Xu Han and Zhiyuan Liu and Maosong Sun},
      year={2025},
      eprint={2410.07145},
      archivePrefix={arXiv},
      primaryClass={cs.CL},
      url={https://arxiv.org/abs/2410.07145}, 
}

@misc{mamba,
      title={Mamba: Linear-Time Sequence Modeling with Selective State Spaces}, 
      author={Albert Gu and Tri Dao},
      year={2024},
      eprint={2312.00752},
      archivePrefix={arXiv},
      primaryClass={cs.LG},
      url={https://arxiv.org/abs/2312.00752}, 
}

@misc{linear-attn,
      title={Transformers are RNNs: Fast Autoregressive Transformers with Linear Attention}, 
      author={Angelos Katharopoulos and Apoorv Vyas and Nikolaos Pappas and François Fleuret},
      year={2020},
      eprint={2006.16236},
      archivePrefix={arXiv},
      primaryClass={cs.LG},
      url={https://arxiv.org/abs/2006.16236}, 
}

@misc{retnet,
      title={Retentive Network: A Successor to Transformer for Large Language Models}, 
      author={Yutao Sun and Li Dong and Shaohan Huang and Shuming Ma and Yuqing Xia and Jilong Xue and Jianyong Wang and Furu Wei},
      year={2023},
      eprint={2307.08621},
      archivePrefix={arXiv},
      primaryClass={cs.CL},
      url={https://arxiv.org/abs/2307.08621}, 
}

@misc{waleffe2024empiricalstudymambabasedlanguage,
      title={An Empirical Study of Mamba-based Language Models}, 
      author={Roger Waleffe and Wonmin Byeon and Duncan Riach and Brandon Norick and Vijay Korthikanti and Tri Dao and Albert Gu and Ali Hatamizadeh and Sudhakar Singh and Deepak Narayanan and Garvit Kulshreshtha and Vartika Singh and Jared Casper and Jan Kautz and Mohammad Shoeybi and Bryan Catanzaro},
      year={2024},
      eprint={2406.07887},
      archivePrefix={arXiv},
      primaryClass={cs.LG},
      url={https://arxiv.org/abs/2406.07887}, 
}

@misc{rwkv7,
      title={RWKV-7 "Goose" with Expressive Dynamic State Evolution}, 
      author={Bo Peng and Ruichong Zhang and Daniel Goldstein and Eric Alcaide and Xingjian Du and Haowen Hou and Jiaju Lin and Jiaxing Liu and Janna Lu and William Merrill and Guangyu Song and Kaifeng Tan and Saiteja Utpala and Nathan Wilce and Johan S. Wind and Tianyi Wu and Daniel Wuttke and Christian Zhou-Zheng},
      year={2025},
      eprint={2503.14456},
      archivePrefix={arXiv},
      primaryClass={cs.CL},
      url={https://arxiv.org/abs/2503.14456}, 
}

@misc{rwkv6,
      title={Eagle and Finch: RWKV with Matrix-Valued States and Dynamic Recurrence}, 
      author={Bo Peng and Daniel Goldstein and Quentin Anthony and Alon Albalak and Eric Alcaide and Stella Biderman and Eugene Cheah and Xingjian Du and Teddy Ferdinan and Haowen Hou and Przemysław Kazienko and Kranthi Kiran GV and Jan Kocoń and Bartłomiej Koptyra and Satyapriya Krishna and Ronald McClelland Jr. and Jiaju Lin and Niklas Muennighoff and Fares Obeid and Atsushi Saito and Guangyu Song and Haoqin Tu and Cahya Wirawan and Stanisław Woźniak and Ruichong Zhang and Bingchen Zhao and Qihang Zhao and Peng Zhou and Jian Zhu and Rui-Jie Zhu},
      year={2024},
      eprint={2404.05892},
      archivePrefix={arXiv},
      primaryClass={cs.CL},
      url={https://arxiv.org/abs/2404.05892}, 
}

@misc{rwkv4,
      title={RWKV: Reinventing RNNs for the Transformer Era}, 
      author={Bo Peng and Eric Alcaide and Quentin Anthony and Alon Albalak and Samuel Arcadinho and Stella Biderman and Huanqi Cao and Xin Cheng and Michael Chung and Matteo Grella and Kranthi Kiran GV and Xuzheng He and Haowen Hou and Jiaju Lin and Przemyslaw Kazienko and Jan Kocon and Jiaming Kong and Bartlomiej Koptyra and Hayden Lau and Krishna Sri Ipsit Mantri and Ferdinand Mom and Atsushi Saito and Guangyu Song and Xiangru Tang and Bolun Wang and Johan S. Wind and Stanislaw Wozniak and Ruichong Zhang and Zhenyuan Zhang and Qihang Zhao and Peng Zhou and Qinghua Zhou and Jian Zhu and Rui-Jie Zhu},
      year={2023},
      eprint={2305.13048},
      archivePrefix={arXiv},
      primaryClass={cs.CL},
      url={https://arxiv.org/abs/2305.13048}, 
}

@misc{lact,
      title={Test-Time Training Done Right}, 
      author={Tianyuan Zhang and Sai Bi and Yicong Hong and Kai Zhang and Fujun Luan and Songlin Yang and Kalyan Sunkavalli and William T. Freeman and Hao Tan},
      year={2025},
      eprint={2505.23884},
      archivePrefix={arXiv},
      primaryClass={cs.LG},
      url={https://arxiv.org/abs/2505.23884}, 
}

@inproceedings{scaling-up-state-size,
    title = "Scaling up the State Size of {RNN} {LLM}s for Long-Context Scenarios",
    author = "Liu, Kai  and
      Gao, Jianfei  and
      Chen, Kai",
    editor = "Che, Wanxiang  and
      Nabende, Joyce  and
      Shutova, Ekaterina  and
      Pilehvar, Mohammad Taher",
    booktitle = "Proceedings of the 63rd Annual Meeting of the Association for Computational Linguistics (Volume 1: Long Papers)",
    month = jul,
    year = "2025",
    address = "Vienna, Austria",
    publisher = "Association for Computational Linguistics",
    url = "https://aclanthology.org/2025.acl-long.564/",
    pages = "11516--11529",
    ISBN = "979-8-89176-251-0"
}

@misc{sse,
      title={Scaling Linear Attention with Sparse State Expansion}, 
      author={Yuqi Pan and Yongqi An and Zheng Li and Yuhong Chou and Ruijie Zhu and Xiaohui Wang and Mingxuan Wang and Jinqiao Wang and Guoqi Li},
      year={2025},
      eprint={2507.16577},
      archivePrefix={arXiv},
      primaryClass={cs.LG},
      url={https://arxiv.org/abs/2507.16577}, 
}

@misc{mom,
      title={MoM: Linear Sequence Modeling with Mixture-of-Memories}, 
      author={Jusen Du and Weigao Sun and Disen Lan and Jiaxi Hu and Yu Cheng},
      year={2025},
      eprint={2502.13685},
      archivePrefix={arXiv},
      primaryClass={cs.CL},
      url={https://arxiv.org/abs/2502.13685}, 
}

@misc{hgrn2,
      title={HGRN2: Gated Linear RNNs with State Expansion}, 
      author={Zhen Qin and Songlin Yang and Weixuan Sun and Xuyang Shen and Dong Li and Weigao Sun and Yiran Zhong},
      year={2024},
      eprint={2404.07904},
      archivePrefix={arXiv},
      primaryClass={cs.CL},
      url={https://arxiv.org/abs/2404.07904}, 
}

@misc{ruler,
      title={RULER: What's the Real Context Size of Your Long-Context Language Models?}, 
      author={Cheng-Ping Hsieh and Simeng Sun and Samuel Kriman and Shantanu Acharya and Dima Rekesh and Fei Jia and Yang Zhang and Boris Ginsburg},
      year={2024},
      eprint={2404.06654},
      archivePrefix={arXiv},
      primaryClass={cs.CL},
      url={https://arxiv.org/abs/2404.06654}, 
}

@misc{infinitebench,
      title={$\infty$Bench: Extending Long Context Evaluation Beyond 100K Tokens}, 
      author={Xinrong Zhang and Yingfa Chen and Shengding Hu and Zihang Xu and Junhao Chen and Moo Khai Hao and Xu Han and Zhen Leng Thai and Shuo Wang and Zhiyuan Liu and Maosong Sun},
      year={2024},
      eprint={2402.13718},
      archivePrefix={arXiv},
      primaryClass={cs.CL},
      url={https://arxiv.org/abs/2402.13718}, 
}

@misc{deltanet,
      title={Linear Transformers Are Secretly Fast Weight Programmers}, 
      author={Imanol Schlag and Kazuki Irie and Jürgen Schmidhuber},
      year={2021},
      eprint={2102.11174},
      archivePrefix={arXiv},
      primaryClass={cs.LG},
      url={https://arxiv.org/abs/2102.11174}, 
}

@misc{gau,
      title={Transformer Quality in Linear Time}, 
      author={Weizhe Hua and Zihang Dai and Hanxiao Liu and Quoc V. Le},
      year={2022},
      eprint={2202.10447},
      archivePrefix={arXiv},
      primaryClass={cs.LG},
      url={https://arxiv.org/abs/2202.10447}, 
}

@misc{eval-harness,
  author       = {Gao, Leo and Tow, Jonathan and Abbasi, Baber and Biderman, Stella and Black, Sid and DiPofi, Anthony and Foster, Charles and Golding, Laurence and Hsu, Jeffrey and Le Noac'h, Alain and Li, Haonan and McDonell, Kyle and Muennighoff, Niklas and Ociepa, Chris and Phang, Jason and Reynolds, Laria and Schoelkopf, Hailey and Skowron, Aviya and Sutawika, Lintang and Tang, Eric and Thite, Anish and Wang, Ben and Wang, Kevin and Zou, Andy},
  title        = {The Language Model Evaluation Harness},
  year         = 2024,
  publisher    = {Zenodo},
  version      = {v0.4.3},
  doi          = {10.5281/zenodo.12608602},
  url          = {https://zenodo.org/records/12608602}
}

@misc{arora2024just,
      title={Just read twice: closing the recall gap for recurrent language models}, 
      author={Simran Arora and Aman Timalsina and Aaryan Singhal and Benjamin Spector and Sabri Eyuboglu and Xinyi Zhao and Ashish Rao and Atri Rudra and Christopher Ré},
      year={2024},
      eprint={2407.05483},
      archivePrefix={arXiv},
      primaryClass={cs.CL},
      url={https://arxiv.org/abs/2407.05483}, 
}

@misc{bick2025understanding,
      title={Understanding the Skill Gap in Recurrent Language Models: The Role of the Gather-and-Aggregate Mechanism}, 
      author={Aviv Bick and Eric Xing and Albert Gu},
      year={2025},
      eprint={2504.18574},
      archivePrefix={arXiv},
      primaryClass={cs.LG},
      url={https://arxiv.org/abs/2504.18574}, 
}

@inproceedings{min-etal-2022-rethinking,
    title = "Rethinking the Role of Demonstrations: What Makes In-Context Learning Work?",
    author = "Min, Sewon  and
      Lyu, Xinxi  and
      Holtzman, Ari  and
      Artetxe, Mikel  and
      Lewis, Mike  and
      Hajishirzi, Hannaneh  and
      Zettlemoyer, Luke",
    editor = "Goldberg, Yoav  and
      Kozareva, Zornitsa  and
      Zhang, Yue",
    booktitle = "Proceedings of the 2022 Conference on Empirical Methods in Natural Language Processing",
    month = dec,
    year = "2022",
    address = "Abu Dhabi, United Arab Emirates",
    publisher = "Association for Computational Linguistics",
    url = "https://aclanthology.org/2022.emnlp-main.759/",
    doi = "10.18653/v1/2022.emnlp-main.759",
    pages = "11048--11064"
}
